\documentclass[twocolumn]{article}

\usepackage{PRIMEarxiv}
\usepackage{authblk}
\usepackage[utf8]{inputenc} 
\usepackage[T1]{fontenc}    
\usepackage{hyperref}       
\usepackage{url}            
\usepackage{booktabs}       
\usepackage{amsfonts}       
\usepackage{nicefrac}       
\usepackage{microtype}      
\usepackage{lipsum}
\usepackage{fancyhdr}       
\usepackage{graphicx}       
\graphicspath{{media/}}     
\usepackage{amsmath}
\usepackage{booktabs} 
\usepackage{subfig}
\usepackage{graphicx}
\usepackage{wrapfig}

\title{ViDAS: Vision-based Danger Assessment and Scoring
\thanks{
\textbf{Authors. Title. Pages.... DOI:000000/11111.}} 
}

\author[1]{Pranav Gupta}
\author[1]{Advith Krishnan}
\author[1]{Naman Nanda}
\author[2]{Ananth Eswar}
\author[1]{Deeksha Agarwal}
\author[1]{Pratham Gohil}
\author[1]{Pratyush Goel}

\affil[1]{SRM Institute of Science and Technology, Chennai, India}
\affil[2]{Vellore Institute of Technology, Vellore, India}

\begin{document}
\twocolumn[{
\begin{center}
\maketitle
\vspace{-3.5em}
\end{center}
}]

\begin{abstract}
We present a novel dataset aimed at advancing danger analysis and assessment by addressing the challenge of quantifying danger in video content and identifying how human-like a Large Language Model (LLM) evaluator is for the same. This is achieved by compiling a collection of 100 YouTube videos featuring various events. Each video is annotated by human participants who provided danger ratings on a scale from 0 (no danger to humans) to 10 (life-threatening), with precise timestamps indicating moments of heightened danger. Additionally, we leverage LLMs to independently assess the danger levels in these videos using video summaries. We introduce Mean Squared Error (MSE) scores for multimodal meta-evaluation of the alignment between human and LLM danger assessments. Our dataset not only contributes a new resource for danger assessment in video content but also demonstrates the potential of LLMs in achieving human-like evaluations.
\end{abstract}

\section{Introduction}
Understanding the danger imminent to a human in a video is a complex task due to shifts in understanding of danger/risk continuously imposed by new knowledge \cite{paltrinieri2019learning}, creating new contexts and definitions. While the meaning of danger itself is subjective (since even expert judgments are based on mental shortcuts, and heuristics, which are susceptible to biases \cite{brito2023subjective}), in a general sense, one can define it as "the likelihood and severity of harm, and the immediacy of a threat". It is necessary to be able to perform effective and automatic danger assessments in videos since they hold immense potential across various fields. Real-time alerts in safety systems, content moderation for online platforms (eg. query-based video searches augmented by automatic danger assessment could consider the danger level displayed in a video and omit it for audiences susceptible to emotional distress), and autonomous systems for navigating hazardous environments are just a few promising applications. However, achieving this remains an ongoing challenge.
\begin{center}
    \captionsetup{type=figure}
    \includegraphics[width=\linewidth]{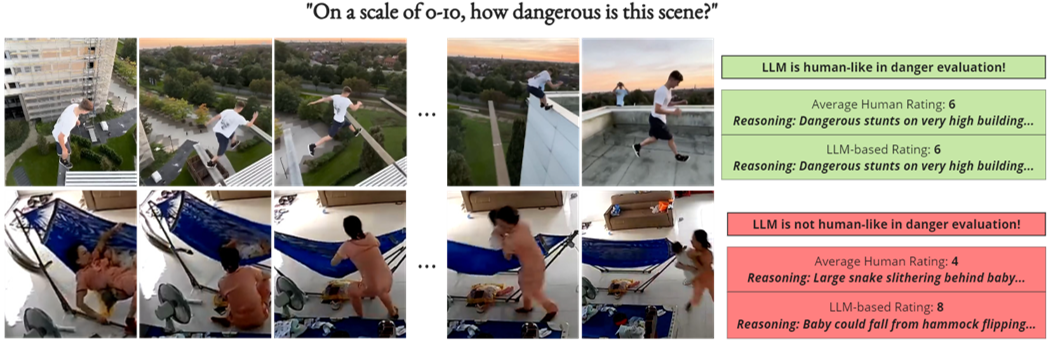}
    \captionof{figure}{Given a video paired with a score assigned by humans on how dangerous the scene is, we evaluate LLMs on how well they can perceive danger via the score it returns. Above are two examples from our dataset.}
\end{center}

There is more focus on "danger detection" methods rather than "danger assessment" methods, which have their limitations when it comes to the generalization of danger context. Many approaches focus on a narrow scope, whether it be a specific method of detection \cite{belsare2024context, wang2023improving, jin2020risk, de2019detection} or specific types of dangers \cite{zhou2022student, wang2023improving, zhu2024deep, li2022pedestrian}, neglecting the broader context crucial for accurate risk assessment. Non-generalized methodologies in temporal vision scenarios can lose sight of the context of danger outside of its domain, object-object relationships, and localized actions across time, which are key aspects when considering a scene to be dangerous to humans present within it. Seeing the vast usage of convolutional network (CNN) models \cite{jean2021study, rajesh2020deep, marchella2023convolutional, wenqi2017model, Roy2024RoadAD, Iyavoo2024PerformanceAO} for danger/risk detection, we often find the requirement of vast amounts of labeled data, which can be time-consuming to collect, but is necessary for building generalized danger detection systems through CNNs.

Understanding the perception of danger in a deep learning methodology can be achieved through analysis of how the methodology assesses the danger in a scene, showcasing that assessment of danger in any given scene can be more informative than the mere detection of danger. Danger assessment through video analysis necessitates pinpointing dangerous elements in a video and gauging the risk level portrayed. This goes beyond mere object recognition and/or classification, requiring an understanding of context, dynamics, and potential consequences enunciated across time. 

LLMs can be very beneficial in fulfilling the highlighted necessities in danger assessors/evaluators due to their few-shot learning capabilities \cite{brown2020language} \& generalizability \cite{lotfi2023non}. The usage of LLMs for video analysis and danger assessment can be considered a promising solution in progressing the field of danger assessment over danger detection. To extend to this, measuring risk and danger through a defined "metric" would assist in defining danger in a quantifiable and objective manner, ranging from utterly zero potential harm to grave, immediate, and life-threatening danger to the human present in a given scenario. LLMs can then assess the dangers shown in input scenes and score it based on this metric, while also allowing them to explain the reasoning behind its score-based assessment through instruction prompts.

Our work presents novel contributions through two key areas. First, we introduce a benchmark dataset for danger assessment in videos. This dataset encompasses a diverse range of danger scenarios, along with annotations capturing the presence of danger, paired with a rubric-based metric system that represents the severity and immediacy shown in the annotated time frames on a scale of 0-10. Second, we leverage this dataset to compare the performance of humans and LLMs in danger assessment tasks. We aim to: 

\textbf{1) Establish a Standardized Benchmark} by providing a common ground for evaluating and comparing future danger assessment models.

\textbf{2) Understand human and LLM danger perception} through exploration of the similarities of danger evaluation via a quantifiable metric.

\textbf{3) Uncover new research directions in danger assessment} through finding areas for improvement and limitations in current approaches in terms of perception of danger, adherence to rubric instructions, etc.

\section{Related Works}

\textbf{Hazardous scene classification}: Hazardous/Dangerous scene classification is the method of automatically identifying and categorizing scenes within images or videos that depict potentially dangerous situations, like workplace accidents, road accidents, riots, assaults, extreme sports, etc.

Consistent research is being done in hazardous, dangerous, and anomalous scene classification and the type of contributions they provide to the field can be broadly split into datasets, showcasing anomalous and harmful scenarios for humans \cite{5539872, mullen2024dontforgetmilkback, 9093457, 8578776} and methodologies proposed for detecting these scenarios apart from normal and harmless scenes, while also classifying them into various categories \cite{5539872, 10.1145/3584376.3584589}.  

Since the boundary between a normal and a dangerous, anomalous scene is often ambiguous \cite{8578776}, evaluating danger based on a pre-defined scale and rubric could help mitigate this issue by extrapolating to a scale of danger levels in increasing order, thereby giving an estimation of its severity to any humans present in the scene.

The importance of this line of research stems from a wide range of benefits that arise by providing meaningful solutions to this task, such as improved public safety, enhanced situational awareness, accident prevention, etc. 

\textbf{Temporal Action Localization}: Temporal Action Localization (TAL) aims to identify the temporal boundaries (start and end times) and spatial regions (bounding boxes) of specific actions within untrimmed videos.
A general overview of the various approaches for TAL pipelines can be listed as follows:

\textit{1) One-stage pipeline:} This directly predicts the start, end time, and bounding box of the action in a single step. (e.g. Convolutional De-Convolutional Networks \cite{8099638}).

\textit{2) Two-stage pipeline:} This approach first generates proposals for potential action locations and then classifies them in a second stage. This allows for more complex reasoning about the video content but can be computationally expensive. (e.g. Background Suppression Networks \cite{Lee_Uh_Byun_2020}).

\textit{3) Anchor-free pipeline:} This eliminates the use of predefined anchors for action proposals, allowing for more flexibility in handling diverse action shapes and sizes. This is a recent development that shows promise for improving localization accuracy. (e.g. Actionness-Guided Transformer \cite{9633209}).

Beyond the pipeline structures, recent research in TAL explores techniques to improve action localization accuracy and efficiency. A variant of this field of research, known as Temporal Action Proposal Generation (TAPG), aims to locate temporal instances of actions in untrimmed videos using the generation of proposals that estimate an action instance's timeframe within a video and evaluate the proposal's prediction through a confidence score \cite{jimaging8080207, yang2021temporalactionproposalgeneration}.

An emerging technique, known as Vision-Language Prompting, leverages natural language descriptions to guide the model in identifying specific actions within videos \cite{nag2022zeroshottemporalactiondetection}. Activity Localization in a video based on a language query by capturing actions in a video as a temporal subgraph consisting of spatial subgraphs for contextualization of the language-conditioned scenes \cite{9423269} is a great example that showcases the potential in Vision-Language Prompting for TAL tasks.

This research area closely aligns with our work on dangerous scene classification, as many dangerous situations involve specific actions (e.g., fighting, falling). Through our work, we try to focus on Vision-Language Prompting for interpreting a given video, localizing actions and movements that take place that seem dangerous, and perceiving the "level of danger" in the presented scene.

\begin{figure*}[t]
	\centering
		\subfloat[]{\includegraphics[width=0.5\linewidth]{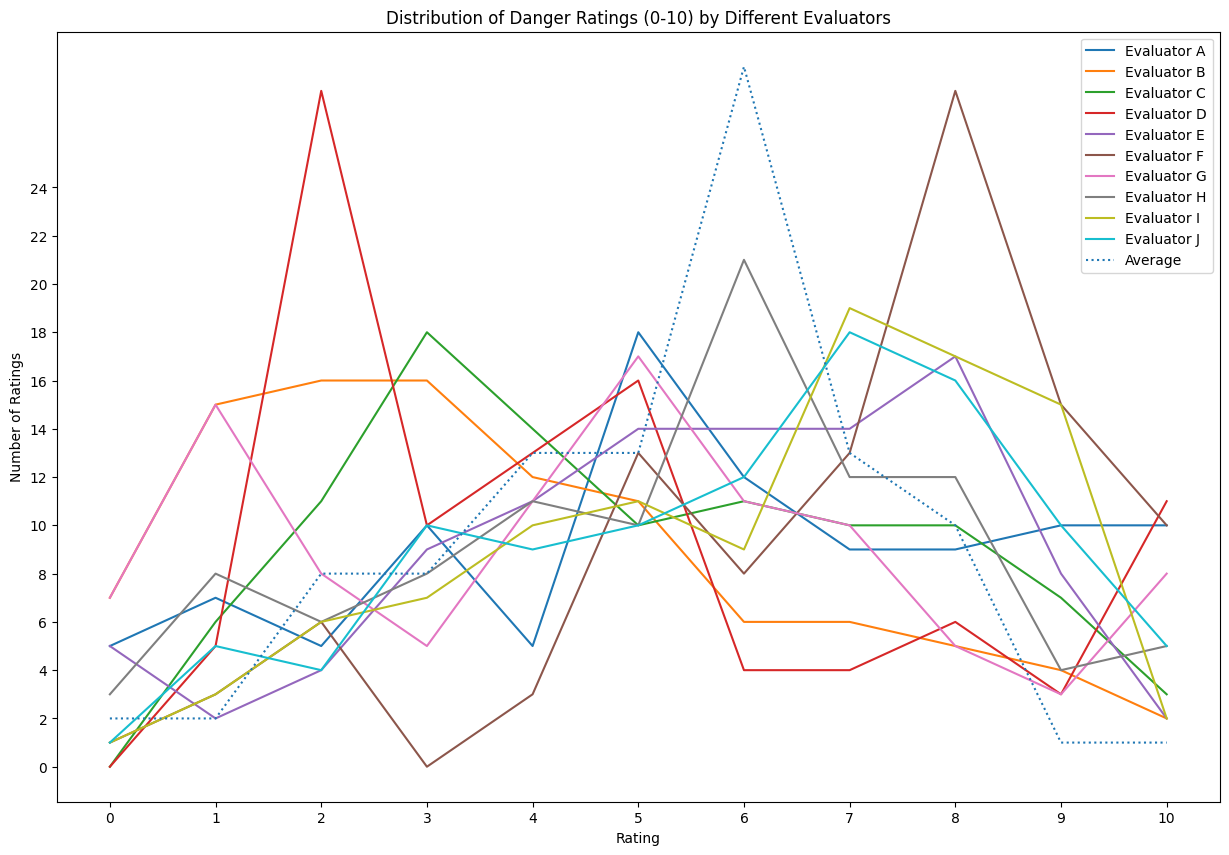}
		\label{fig:ratingdist}}
		\subfloat[]{\includegraphics[width=0.5\linewidth]{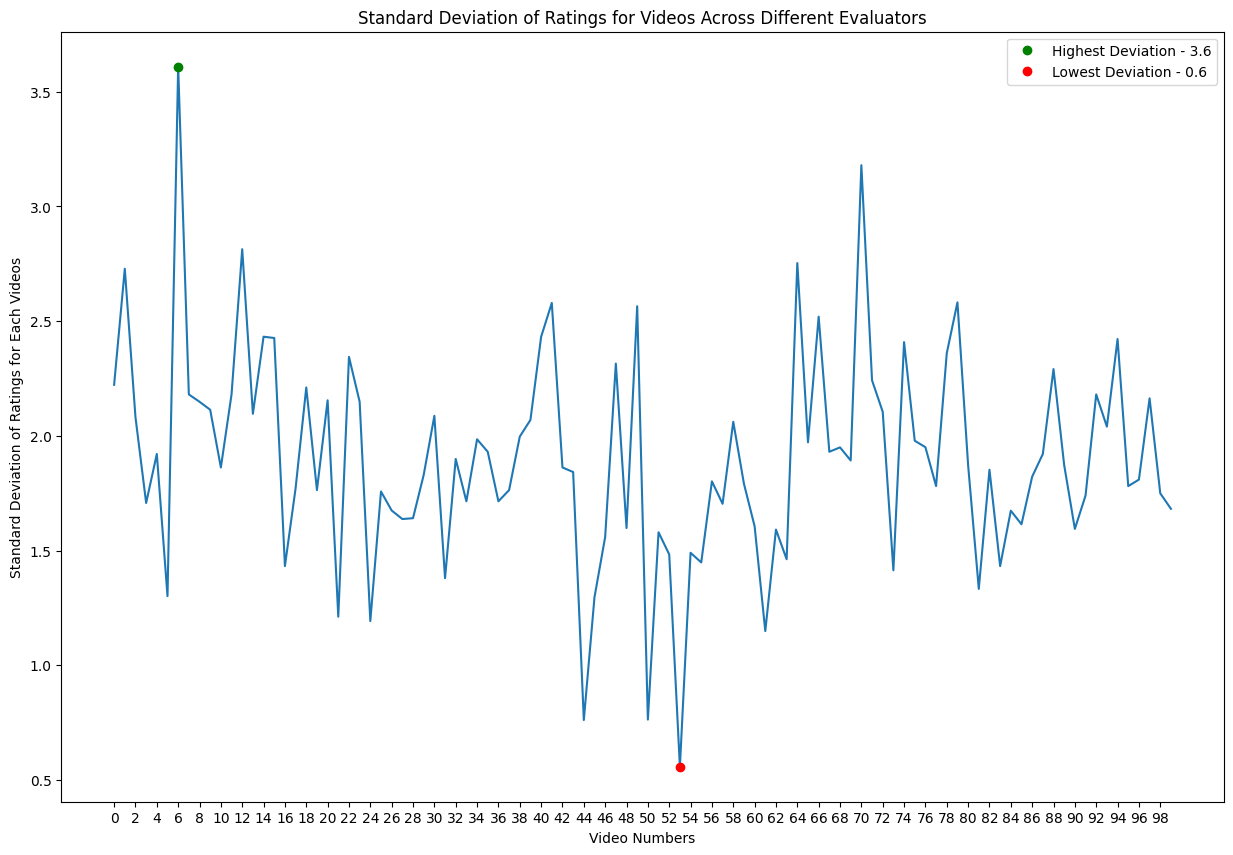}
		\label{fig:ratingdev}}
	\caption{Plots show (a) distribution of 10 random evaluators' ratings and each video's average ratings. (b) describes the standard deviation of the seven evaluators for all 100 videos}
	\label{fig:curves}

\end{figure*}

\textbf{Zero-shot \& Few-shot Instruction}: Zero-shot Instruction (ZSI) and Few-shot Instruction (FSI) are two learning paradigms for LLMs where the model can perform a new task with minimal or no training data specific to that task, through prompt engineering and fine-tuning that helps to instruct how the LLM must behave for specific use-cases. In-context learning using LLMs has been an interesting topic for research lately due to its performance benchmarks and several recent studies have explored the effectiveness of few-shot instructions in guiding LLMs toward specific tasks \cite{brown2020language, coda2023meta, Li_2023, prabhumoye2022fewshotinstructionpromptspretrained}. The potential of concise instructions to improve LLM performance for various LLM tasks is demonstrated in the GPTscore evaluation framework which utilizes zero-shot instructions on generative pre-trained models \cite{fu2023gptscoreevaluatedesire}. 

Apart from that, pinpointing specific moments within a video based on natural language queries, also known as Moment Localization \cite{gao2017talltemporalactivitylocalization, hendricks2017localizingmomentsvideonatural, zhang2019learningsparse2dtemporal}, is a noteworthy use-case of zero-shot/few-shot instructions, leaving ample room for discussion of its capabilities in our task. Emphasis on the importance of understanding the relationships between objects in a scene to enhance moment localization accuracy is another research field that can be associated with our task, as incorporating language-conditioned graph learning into a zero-shot or few-shot framework might enable LLMs to generalize better to unseen query types and improve localization performance.

Our work extends this line of research by investigating how zero-shot and few-shot instructions can guide LLMs toward danger assessment based on a pre-defined rubric, specifically within the context of vision-language inputs like CCTV footage, camera footage, etc.

\textbf{LLM-based Evaluators \& Meta-Evaluation}: 
LLM-based evaluators are LLMs specifically designed to assess the quality and performance of other LLMs. By automatically analyzing LLM generations, these evaluators provide quantitative and qualitative feedback on how well an LLM can perform a certain task, enabling the ease of evaluation. This automation offers significant advantages over traditional human/manual evaluation methods in terms of speed and scalability.

Meta-evaluation refers to the process of evaluating the evaluation methods themselves. It involves assessing the reliability and validity of an LLM-based evaluation method through metrics, benchmarks, and even other LLMs instead of human judgment used to measure model performance. Datasets that are used as benchmarks to evaluate LLM evaluators themselves are Meta-Evaluation Benchmarks (MEBs).

Creating robust MEBs for scenarios that utilize evaluation by LLMs remains an active and unexplored area of research for multimodal tasks like language-conditioned video analysis. Existing work not only showcases the potential of LLMs for automated evaluation \cite{lin2023llmevalunifiedmultidimensionalautomatic, liu2023calibratingllmbasedevaluator} but also demonstrates the importance of standardized datasets for multi-task evaluation \cite{wang2019gluemultitaskbenchmarkanalysis} as well as better human alignment in terms of the assessment of summarization, data-to-text, and hallucinations \cite{liu2023calibratingllmbasedevaluator}. 

Our work complements existing research by proposing a new benchmark for evaluation within a multimodal context, combining a rubric-based instruction prompt meant for a spatiotemporal input like hazardous/dangerous scene footage. This helps us understand the capabilities of vision-language models outside of traditional text-only tasks through evaluation by other LLMs, which we achieve through multimodal meta-evaluation designed to assess MLLM performance in danger assessment, which is a very ambiguous task, through explanation by a metric that aligns with human-level understanding.

\begin{figure*}[t]
\centering
    \subfloat[]{\includegraphics[width=0.5\linewidth]{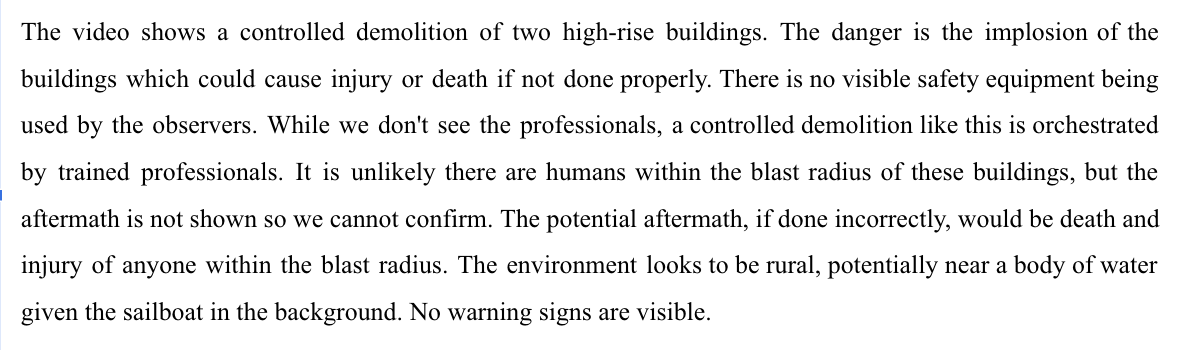}
    \label{fig:6}}
    \subfloat[]{\includegraphics[width=0.4\linewidth]{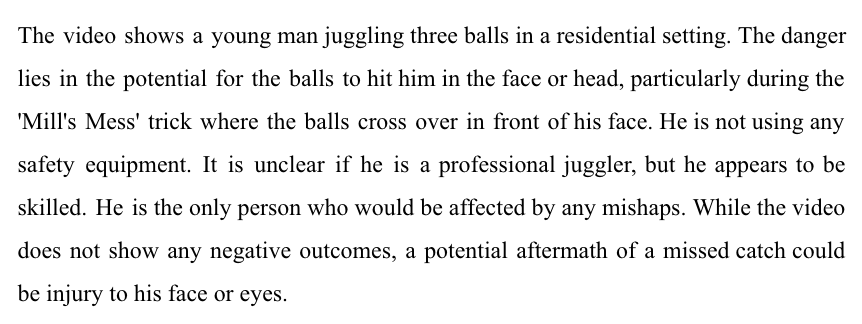}
    \label{fig:53}}
\caption{Example video summaries. Video (a) is given an average rating of \( E_a^{(\text{avg})} \)=5 and (b) is given an average rating of \( E_b^{(\text{avg})} \)=0}
\label{fig:summaries}

\end{figure*}

\section{Dataset Details}
\subsection{Structure}
Our research proposes a novel danger metric for video content, ranging from 0 (no danger) to 10 (extreme danger). To establish a clear frame of reference for applying this metric, a structured framework categorizes potential dangers into four key areas: extreme sports, accidents, stunts, and workplace hazards/natural disasters.

The collected video data encompasses a broad spectrum of risk levels within each category. Danger levels are assigned based on the inherent relative risk to humans present in the scenario depicted.

For example, beginner-level rock climbing receives a danger level of 1, reflecting a low inherent risk, while wingsuit flying near the ground is categorized as a danger level 10 due to the extreme potential for serious injury or death. Similarly, danger levels are assigned within workplace settings, ranging from minor slips (level 4) to major building collapses (level 10). Highlighting the timestamp of the video segment containing the danger component allows for a more precise understanding of relative risk within the established 0-10 scale.

This structured framework not only defines the danger metric but also demonstrates its versatility across diverse scenarios, paving the way for potential applications in areas such as content moderation, and safety assessment.
\subsection{Data Collection}\label{sec:data-collect}
A curated manual selection of YouTube videos featuring danger of varying levels and various events was used to gather data for this research.

Pytube was utilized, which is a lightweight library written in Python that has no third-party dependencies, for the installation of the selected videos. 
Metadata was further created, which comprised information about each video, including the Video ID, Title, Description, URL, Channel Name, Duration, and a temporary filename for reference of each video.

\subsection{Annotation Pipeline}\label{sec:annotation-pipeline}
The VGG Video Annotator facilitates an efficient video annotation process through systematic steps. This tool was used so that humans could identify temporal segments and edit the timestamp to focus on specific parts of the video that contain the danger component. 

The danger metric, indicating the severity of danger, can be assigned to the video on a scale of 0-10. The annotated project can be pushed and saved to a server, and users are given the necessary functionalities to navigate through the video dataset, and evaluate and mark the danger in each video presented as shown in Figures 1 \& 2.

This structured workflow ensures accurate and systematic annotation of videos, enabling thorough and unbiased contextual analysis and eliminating any contextual intervention that may arise while navigating through the dataset.
\begin{figure}
    \centering
    \includegraphics[width=1.05\linewidth]{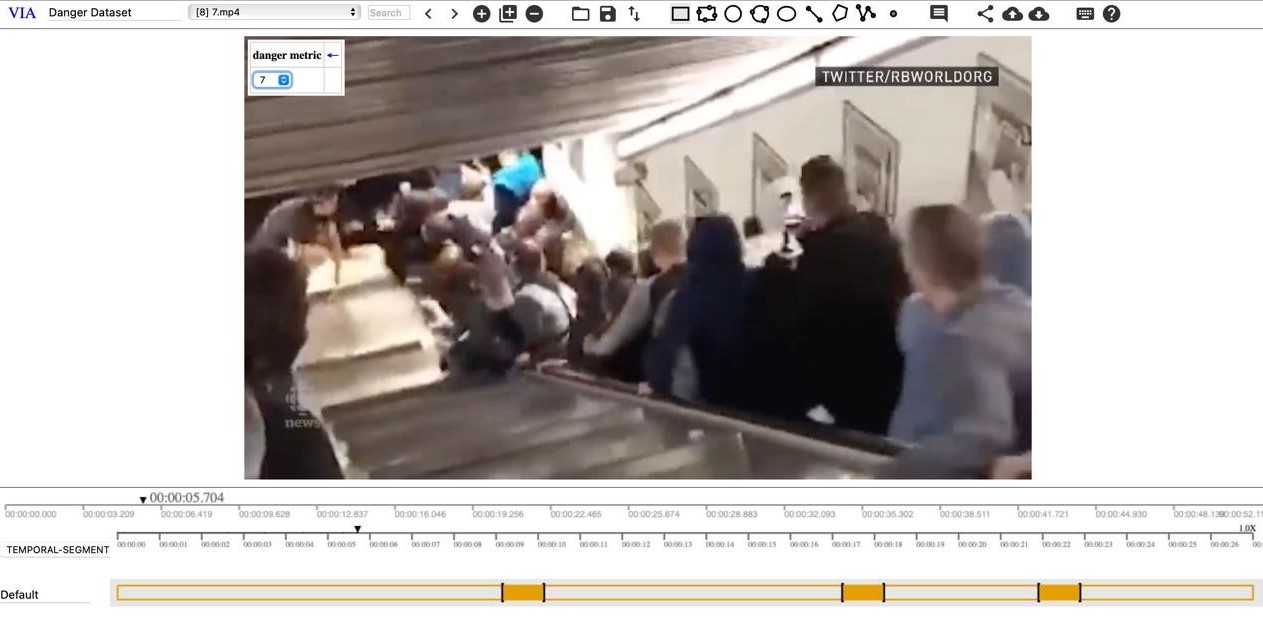}
    \caption{Marking the danger rating and timeframes of heightened danger using VGG Video Annotator}
    \label{fig:annotator}
\end{figure}

\begin{figure}
    \centering
    \includegraphics[width=1.05\linewidth]{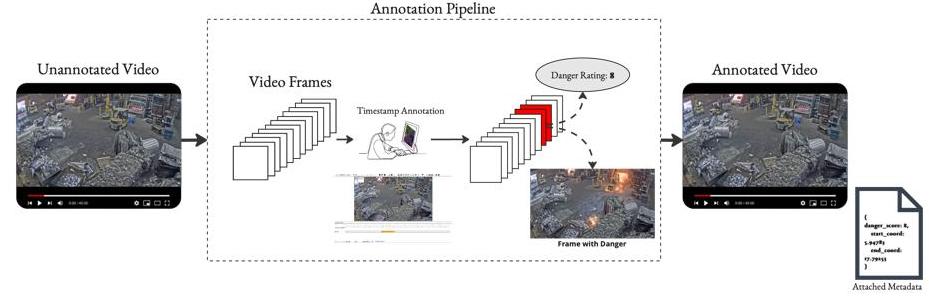}
    \caption{Human Annotation Pipeline}
    \label{fig:pipeline}
\end{figure}

\section{Dataset Statistics}
We present a detailed analysis of the dataset used for our study, focusing on various statistical measures from the evaluations of seven human participants.

Through the process detailed in Section \ref{sec:data-collect} we collected 100 videos with varying danger levels and scenarios. Using the Annotation Pipeline discussed in Section \ref{sec:annotation-pipeline} we collected 18 human responses.

\textbf{Danger Rating:} Each participant assigned a rating that quantifies the perceived danger level of the event depicted in each video. Let \( E_i^{(j)} \) represent the danger rating given by the \( j \)-th evaluator for the \( i \)-th video.

The average evaluator rating for the \( i \)-th video, denoted as \( E_i^{(\text{avg})} \), is calculated by averaging the ratings given by all evaluators. If there are \( n \) evaluators, the average rating can be computed as follows:

\[
E_i^{(\text{avg})} = \frac{1}{n} \sum_{j=1}^{n} E_i^{(j)}
\]

\textbf{Temporal Timestamps:} Participants identified and annotated the start and end points of the event within the video, providing precise temporal coordinates for the duration of the dangerous activity. We calculate the average ratings and temporal coordinates from the \( n \) evaluations for each video, establishing these averages as the ground truth temporal segments. Let \( T_{i,\text{start}}^{(j)} \) and \( T_{i,\text{end}}^{(j)} \) denote the start and end points annotated by the \( j \)-th evaluator for the \( i \)-th video, respectively. The average start and end points for the \( i \)-th video can be computed as follows:

\[
T_{i,\text{start}}^{(\text{avg})} = \frac{1}{n} \sum_{j=1}^{n} T_{i,\text{start}}^{(j)} \quad T_{i,\text{end}}^{(\text{avg})} = \frac{1}{n} \sum_{j=1}^{n} T_{i,\text{end}}^{(j)}
\]
These average temporal coordinates are used as the ground truth temporal segments for each video.

\textbf{Distribution of Ratings:}
The variability in danger ratings among different evaluators is evident in Figure \ref{fig:ratingdist}, indicating differing perceptions of danger. The average rating distribution, denoted by the dotted line, shows the distribution of the ground truth labels. Evaluator A exhibits almost a similar frequency of all danger ratings, whereas Evaluator D consistently assigns lower ratings. Evaluator F in contrast, has rated more higher ratings to videos.

\textbf{Deviation of Ratings for Videos Across Different Evaluators:}
Videos with high standard deviation scores as shown in Figure \ref{fig:ratingdev} indicate significant disagreement among evaluators. This suggests that different individuals perceive the danger levels of these videos very differently. For example, Figure \ref{fig:6} discusses a controlled demolition. While some perceived that this demolition was conducted without any humans around it (lower scores), others felt it was still a dangerous event. 

Conversely, videos with low standard deviation scores indicate a strong consensus among evaluators about the danger level. These videos are typically more straightforward in terms of the danger presented, with clear risks that are easily identifiable. For example, a video showed a boy juggling three balls; the only potential danger of this event is the balls dropping on him. Almost all evaluators rated this a 0 as shown in Table \ref{tab:ratingdiff}.

\begin{figure}
    \centering
    \includegraphics[width=\linewidth]{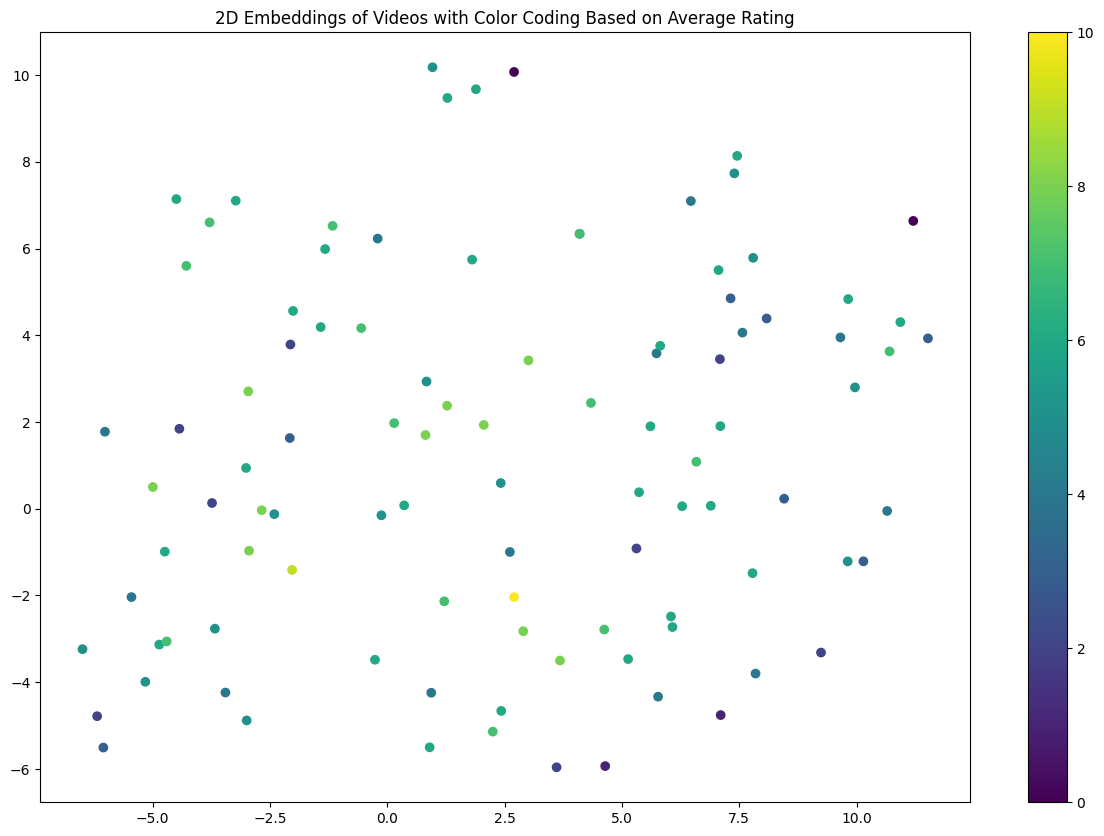}
    \caption{Marking the danger rating and timeframes of heightened danger using VGG Video Annotator}
    \label{fig:tsne}
\end{figure}

\textbf{Spread of the Video Summaries:}
The t-SNE graph in Figure \ref{fig:tsne} shows a 2D embedding of video summaries, with each point representing a video. The videos are color-coded based on their average danger ratings, as indicated by the color bar on the right. The x and y axes represent the two-dimensional embeddings of the videos obtained through a dimensionality reduction. The spread of points across the plot suggests a diverse distribution of video summaries. The color gradient from dark blue (low danger rating) to yellow (high danger rating) provides a visual representation of how different videos are perceived in terms of danger.

\begin{table}[h]
    \centering
    \begin{tabular}{{|p{0.6cm}|p{0.35cm}|p{0.35cm}|p{0.35cm}|p{0.35cm}|p{0.35cm}|p{0.35cm}|p{0.35cm}|p{0.35cm}|p{0.35cm}|p{0.35cm}|}}
        \hline
        Video & \( E_i^{(\text{0})} \) & \( E_i^{(\text{1})} \) & \( E_i^{(\text{2})} \) & \( E_i^{(\text{3})} \) & \( E_i^{(\text{4})} \) & \( E_i^{(\text{5})} \) & \( E_i^{(\text{6})} \) & \( E_i^{(\text{7})} \) & \( E_i^{(\text{8})} \) & \( E_i^{(\text{9})} \)\\
        \hline
        \ref{fig:6} & 5 & 0 & 9 & 9 & 7 & 4 & 10 & 5 & 8 & 2 \\
        \hline
        \ref{fig:53} & 0 & 0 & 1 & 2 & 0 & 0 & 0 & 0 & 0 & 0 \\
        \hline
    \end{tabular}
    \caption{Table shows 10 random evaluations of 2 videos rated with a consensus and different perceptions of danger. }
    \label{tab:ratingdiff}
\end{table}
\section{Methodology}\label{sec:method}

\begin{figure*}[t]
    \centering
    \subfloat[]{\includegraphics[width=1\linewidth]{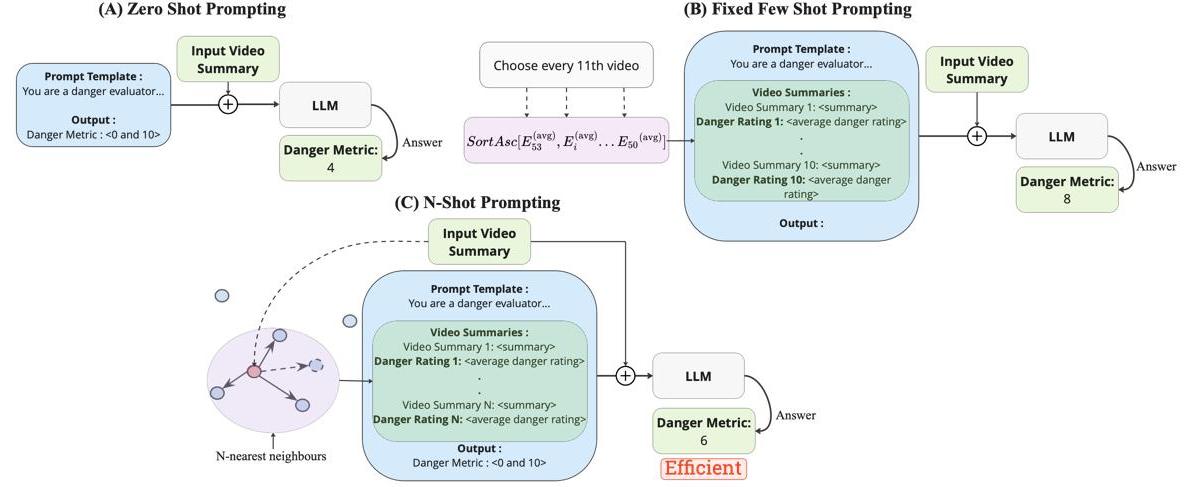}}
    \caption{This figure illustrates three prompting methods for evaluating danger levels in video summaries with Large Language Models (LLMs). (A) Zero-Shot Prompting uses a template and video summary without prior examples. (B) Fixed Few Shot Prompting selects every 11th video summary as predefined examples based on sorted ratings. (C) N-Shot Prompting dynamically selects relevant examples based on similarity to the target summary.}
    \label{fig:prompting-methodologies}
\end{figure*}

This section details the methodology employed using large language models (LLMs) to assess danger in video content. Our research involves some key approaches like Video Summarization where we use LVMs to summarize each video based on the key dangerous events being shown. 

We then utilize different LLMs and prompting techniques as shown in Figure \ref{fig:prompting-methodologies} to guide and retrieve danger metric evaluations on each video in the dataset from the LLM. 

\subsection{Video Summarization}\label{video-summarization}
Video summarization is a crucial step in our methodology as it condenses extensive video content into concise and informative summaries. The summaries highlight each video's key events and actions, particularly those that contribute to the perceived danger.

We leveraged Large Vision Models (LVMs) and their multimodal capabilities in this task. We provide each of the 100 videos as input to the LVM. The model generates a summary for each video, encapsulating the main events and highlighting any dangerous activities.

The following is our prompt template to generate the video summaries given the video as input:

\begin{verbatim}
You will be given a video which contains 
some danger. 
You need to create a summarization of the 
danger-based on this questionnaire:

What is the danger in the video?
Is there any safety equipment being used?
Is a professional working on the gear/danger 
apparatus?
Are there any humans that will be affected 
by it?
If the aftermath was shown of the danger 
event, did anyone die/escape/get injured?
If the aftermath was not shown, what would 
be the potential aftermath

The summary should just be one **short** 
paragraph detailing a description of what 
is happening in the video and the above 
questionnaire.
\end{verbatim}

\subsection{Zero-shot Prompting}\label{sec:zero-shot}
Zero-shot prompting (ZS) is a method where the model performs a specific evaluation without any prior examples or additional context to guide its response. In the context of assessing danger in video content, zero-shot prompting leverages the inherent capabilities of the LLM to understand and evaluate the summarized video content directly.

The primary objective of zero-shot prompting is to obtain an initial danger rating for each video summary without relying on any pre-existing examples. This approach tests the model’s ability to generalize and assess danger based solely on the input prompt and video summary provided.

In the zero-shot prompting approach, each video summary generated from Section \ref{video-summarization} is evaluated using a fixed prompt. This prompt is designed to instruct the model to rate the danger level of the video from 0-10. 

The following is the zero-shot prompt template:
\begin{verbatim}
You are a danger evaluator. Analyze the 
given summary and provide a rating based 
on the analysis of the video summary between 
0 and 10 where 10 is the most dangerous and 
0 is the least. However, keep in mind that 
the danger must come from a situation in 
which a human or a living being may get hurt 
because of lack of familiarity with the 
situation. Don't give any explanations.

Output:
Danger Metric: <A number between 0 and 10>
\end{verbatim}

\begin{table}[h]
    \centering
    \begin{tabular}{{p{2.5cm}cc}}
        \hline
        LLM & Fixed Few shot & Zero-shot \\
        \hline
        gpt-3.5-turbo & 8.84 & 13.50\\
        gpt-4-turbo & \textbf{4.70} & 12.59 \\
        gpt-4o & 5.59 & 12.16\\
        gemini-1.5-pro & 11.01 & \textbf{11.32} \\
        \hline
    \end{tabular}
    \caption{Table shows different Mean Squared Error (MSE) Scores of different LLMs using Fixed Few-Shot and Zero-Shot Prompting.}
    \label{tab:fixed-few-shot-comparison}
\end{table}

\subsection{Fixed Few-shot Prompting}\label{sec:ffs}
Fixed few-shot (FFS) prompting is a method where the model is provided with a set of predefined examples (few shots) along with their annotations to guide its evaluation process.

The main objective of FFS prompting is to enhance the model’s ability to assess danger levels in videos by providing it with a subset of annotated examples. These examples help the model better understand the nuances of danger ratings, leading to more accurate assessments for the remaining videos.

In this approach, we sorted the dataset based on average human-evaluated danger ratings and then selected every 11th video from this sorted list to serve as few-shot examples. These examples, along with their summaries and danger ratings, were used in a prompt template given as context to the LLM.

The following is our FFS prompt template:
\begin{verbatim}
You are a danger evaluator. Analyze the 
given summary and provide a rating based 
on the analysis of the video summary between 
0 and 10 where 10 is the most dangerous and 
0 is the least. However, keep in mind that 
the danger must come from a situation in 
which a human or a living being may get hurt 
because of lack of familiarity with the 
situation. Don't give any explanations.

Video Summary 1: <video summary>
Danger Rating 1: <average danger rating>
.
.
Video Summary 10: <video summary>
Danger Rating 10: <average danger rating>
\end{verbatim}

\subsection{N-shot Prompting}\label{sec:n-shot}
We use N-shot prompting that dynamically selects the most relevant examples for each evaluation task based on their similarity to the target video summary. This approach aims to provide the model with the most contextually appropriate examples, thereby enhancing its ability to assess danger levels accurately.

For each video summary, we first generate its embeddings using a sentence transformer. We compute the cosine similarity scores of each embedding and retrieve the top N summaries based on the highest similarity scores.

The top N summaries are used as few-shot examples along with their human-evaluated danger ratings in a prompt template given to the LLM as context.

The following is our N-shot prompt template:

\begin{table*}[t]
    \centering
    \begin{tabular}{{cccccccccccccccc}}
        \hline
        Model & 1 & 2 & 3 & 4 & 5 & 6 & 7 & 8 & 9 & 10 & 20 & 25 & 30 & 40 & 50 \\
        \hline
        gpt-4o & 9.25 & 8.62 & 8.14 & 7.63 & 7.37 & 7.17 & 6.92 & 6.73 & 6.55 & 6.42 & 4.83 & 4.67 & 4.68 & \textbf{4.31} & 4.83 \\
        gemini-1.5-pro & 8.93 & 8.21 & 7.93 & 7.65 & 7.50 & 7.20 & 7.10 & 7.00 & 6.99 & 6.92 & 6.30 & 6.17 & 6.03 & \textbf{5.82} & 5.95 \\
        \hline
    \end{tabular}
    \caption{Ablation Study of setting different values of N for N-shot prompting}
    \label{tab:N-shot-comparison}
\end{table*}

\begin{verbatim}
You are a danger evaluator. Analyze the 
given summary and provide a rating based 
on the analysis of the video summary between 
0 and 10 where 10 is the most dangerous and 
0 is the least. However, keep in mind that 
the danger must come from a situation in 
which a human or a living being may get hurt 
because of lack of familiarity with the 
situation. Don't give any explanations.

Video Summary 1: <video summary>
Danger Rating 1: <average danger rating>
.
.
Video Summary N: <video summary>
Danger Rating N: <average danger rating>
\end{verbatim}

\section{Results and Discussion}

It is essential to ensure that the LLM can do the following things to effectively perform the task of danger level estimation:
\begin{itemize}
    \item Identify Danger Elements: Accurately detect potential sources of danger in the videos.
    \item Understand Context: Comprehend the context surrounding the danger to make informed assessments.
    \item Gauge Danger Levels: Evaluate the severity of the danger based on appropriate scales and criteria.
\end{itemize}

We experimented with Video Summarization and the prompting techniques detailed in Section \ref{sec:method} with LLMs.

We used Gemini-1.5-Pro as the LVM and fed the prompt from Section \ref{video-summarization} to generate video summaries of all 100 videos. Some examples are shown in Figure \ref{fig:summaries}.

We use OpenAI's text-embedding-3-small sentence transformer to generate embeddings for all video summaries to perform N-shot learning detailed in Section \ref{sec:n-shot}.

We see some discrepancies when providing videos in which the LVM fails to detect a snake behind a hammock with a baby in a particular video. It was only able to detect the snake when it was specifically pointed out.

In all the experiments, we set the temperature to 0 to minimize the variability in the LLM responses.

To compare the danger ratings predicted by a Language Model (LLM) to the average danger ratings given by humans, we use the Mean Squared Error (MSE) as our metric. The MSE provides a measure of the average squared difference between the predicted ratings and the actual average ratings. Let \( L_i \) represent the danger rating predicted by the LLM for the \( i \)-th video, and let \( E_i^{(\text{avg})} \) denote the average danger rating given by human evaluators for the \( i \)-th video. The MSE can be formulated as follows:

\[
\text{MSE} = \frac{1}{n} \sum_{i=1}^{n} (L_i - E_i^{(\text{avg})})^2
\]

where \( n \) is the total number of videos. The MSE thus quantifies the prediction accuracy by averaging the squared differences between the predicted and actual ratings overall videos.

\paragraph{Comparison with different models}
We use the ZS and FFS Prompting approaches with a set prompt template detailed in Figure \ref{fig:prompting-methodologies} and Sections \ref{sec:zero-shot} and \ref{sec:ffs}. From Table \ref{tab:fixed-few-shot-comparison} we see that gpt-4-turbo gets the lowest MSE score in FFS prompting while gpt-3.5-turbo produces the highest. In ZS prompting, gemini-1.5-pro produces the lowest MSE Scores in contrast with gpt-3.5-turbo with the highest. This reflects that newer and bigger LLMs like gpt-4-turbo and gemini-1.5-pro can assess danger closer to humans than smaller LLMs.

\paragraph{Ablation Study}
We also experimented with different values of N using the N-shot prompting approach shown in Table \ref{tab:N-shot-comparison} with two different LLMs: gpt-4o and gemini-1.5-pro.

We see a steady decrease in MSE in both models as the value of N increases, with the lowest score when the value of N is 40 in both models. This infers that by giving more context to the LLM as few-shot examples, it performs better. However, by comparing 10-shot from Table \ref{tab:N-shot-comparison} to the fixed 10-shot from Table \ref{tab:fixed-few-shot-comparison} of gpt-4o we see that performance may not necessarily depend on the size of N but also on the type of examples provided in the prompt.

\section{Limitations and Future Work}
While our research into the assessment of danger by Large Language Models (LLMs) compared to human evaluation has yielded valuable insights, there are several areas where improvements and further work could enhance the robustness and comprehensiveness of our findings.

\textbf{Dataset Size:}
The most notable limitation of a benchmark is defined by the expanse of features and variety of contexts in the data, as well as how well it is generalized for its use case. Our benchmark dataset contains 100 videos, which, while informative, comprise a relatively small sample size that isn't representative of a generalized idea of danger.
Expanding the dataset to include a larger number of videos would solidify the statistical truth of our comparisons and provide a more diverse range of scenarios to discern the behavior and the generalized perception of danger in LLMs.

\textbf{Number of Human Evaluators:}
The number of human evaluators involved in the assessment is limited.
Increasing the number of human evaluators would lead to a more robust and reliable comparison. A larger panel of evaluators would provide a broader perspective on danger assessment, capturing a wider range of human judgment nuances.

\textbf{Occlusions in Videos:}
Objects that are hidden or partially visible (occlusions) may not be adequately assessed by LLMs, potentially leading to inaccuracies in danger evaluation. Implementing advanced computer vision techniques that can handle occlusions and provide a more comprehensive analysis of the scene could enhance the LLM's ability to evaluate danger accurately. Some considerable methods that can help solve occlusion problems could be novel view synthesis methods like 3D Gaussian Splatting and Neural Radiance Fields.

\section{Conclusion}
We introduced a novel dataset for advancing danger analysis and assessment in video content, comprising 100 YouTube videos annotated with precise danger ratings and temporal ratings by human participants. This dataset quantifies danger and evaluates the alignment between human and Large Language Model (LLM) assessments using Mean Squared Error (MSE) scores. Emphasis was placed on capturing diverse danger scenarios with rich annotations, enhancing its utility for developing and benchmarking danger assessment models.

Future research directions include expanding the dataset size and variety of scenarios to improve model robustness and generalizability. Increasing the number of human evaluators will provide a richer understanding of danger perception. Leveraging advanced computer vision techniques, such as 3D Gaussian Splatting and Neural Radiance Fields, could enhance LLMs’ accuracy in evaluating danger.

Our findings highlight the potential of LLMs in achieving human-like danger evaluations and uncover areas for improvement. We are exploring dataset expansion and integration of additional annotations to further drive the development of robust danger assessment systems. This dataset offers a valuable benchmark for future research, contributing to safer and more intelligent AI applications.

For more information and to download the dataset, please visit the supplementary materials.

\textbf{Acknowledgments} We thank all participants and community members for their valuable feedback throughout this project’s development.

\bibliographystyle{unsrt}  
\bibliography{references}

\end{document}